ADEMA: A Knowledge-State Orchestration Architecture for Long-Horizon Knowledge Synthesis with LLM Agents

Zhou Hanlin[1],[2]; Chan Huah Yong[1]*

[1] School of Computer Sciences, Universiti Sains Malaysia (USM), Penang, Malaysia

[2] Xiamen Software Vocational & Technical College, Xiamen, China

* Corresponding author
Email: hychan@usm.my

Abstract

Long-horizon LLM tasks often fail not because a single answer is unattainable, but because knowledge states drift across rounds, intermediate commitments remain implicit, and interruption fractures the evolving evidence chain. This paper presents ADEMA as a knowledge-state orchestration architecture for long-horizon knowledge synthesis rather than as a generic multi-agent runtime. The architecture combines explicit epistemic bookkeeping, heterogeneous dual-evaluator governance, adaptive task-mode switching, reputation-shaped resource allocation, checkpoint-resumable persistence, segment-level memory condensation, artifact-first assembly, and final-validity checking with safe fallback. Evidence is drawn entirely from existing materials: a four-scenario showcase package, a fixed 60-run mechanism matrix, targeted micro-ablation and artifact-chain supplements, and a repaired protocol-level benchmark in which code-oriented evaluation is the clearest quality-sensitive mechanism block. Across the fixed matrix, removing checkpoint/resume produced the only invalid run, and it did so in the interruption-sensitive resume condition. By contrast, dual evaluation, segment synthesis, and dynamic governance are best interpreted as supporting control mechanisms that shape trajectory discipline, explicit artifact progression, and cost-quality behavior rather than as universal binary prerequisites for completion. The contribution is therefore a knowledge-state orchestration architecture in which explicit epistemic state transition, evidence-bearing artifact progression, and recoverable continuity are the primary design commitments.

1. Introduction

Large language models (LLMs) are increasingly deployed as components in compound knowledge systems rather than as isolated chat interfaces. In that setting, the central engineering problem is not merely how to elicit one strong answer, but how to govern knowledge state across rounds, interruptions, intermediate artifacts, and shifting control signals[1, 2].

Recent multi-agent frameworks have demonstrated the value of role specialization, conversational decomposition, and tool mediation for complex tasks [3-5]. Structured-output and long-context studies further show that external constraints, observability, and compression strategies materially affect system behavior[6] .

Yet much of that literature still evaluates success through final-answer completion, framework flexibility, or benchmark throughput. For long-horizon knowledge synthesis, those criteria are incomplete because they do not directly reveal whether the evolving knowledge state remains governable, inspectable, and recoverabl[7].

This paper studies a concrete architecture, ADEMA, that addresses those gaps through eight interacting mechanisms: explicit epistemic bookkeeping, heterogeneous dual-evaluator governance, adaptive task-mode switching, dynamic reputation-based budget allocation, checkpoint-resumable persistence, segment-level memory condensation, artifact-first deliverable assembly, and final-validity checking with safe fallback[8]. These mechanisms are implemented in the reviewed source code as concrete runtime objects rather than as narrative abstractions[9].

The resulting contribution is best understood as a knowledge-state orchestration architecture rather than a conversational workflow. Instead of letting long-horizon synthesis unfold as unconstrained message accumulation, ADEMA externalizes hypothesis state, milestone status, condensed memory, artifact lineage, and resumable checkpoints as explicit system state[10]. In this sense, the architecture's principal value is not generic collaboration superiority, but making hidden long-horizon knowledge progression governable.

The paper makes four contributions. First, it presents ADEMA as a knowledge-state orchestration architecture for governable long-horizon knowledge synthesis. Second, it maps concrete source-code mechanisms to explicit epistemic progression, evidence-bearing artifact progression, and recoverable continuity. Third, it reorganizes the existing experimental materials so that the clearest headline evidence is explicit: checkpoint persistence is the only completion-critical mechanism under interruption, and the code-oriented benchmark is the strongest quality-sensitive mechanism block. Fourth, it provides a reviewer-facing package in which the main manuscript, a single supplementary document, and a single source-code-and-artifacts archive support inspection without unrestricted public release of the full operational environment.

## 2. Related work

### 2.1 LLM multi-agent collaboration and long-horizon task solving

Multi-agent LLM frameworks such as CAMEL, MetaGPT, AutoGen, AgentVerse, and related autonomous-agent systems focus on role specialization, decomposition, and collaboration patterns [11-13]. ReAct and Self-Refine further show that iterative reasoning and feedback can improve solution trajectories even without a full multi-agent setting [14]. These lines of work establish that complex tasks benefit from intermediate reasoning structure rather than one-shot prompting alone.

Yet many shared-thread and group-chat style orchestration patterns still leave the evolving knowledge state implicit. Messages accumulate, but the status of hypotheses, milestones, partially validated conclusions, or artifact readiness is not always modeled as explicit system state. ADEMA targets a different runtime objective: selective state propagation and artifact-bound progression rather than persistent full-context broadcast to all agents.

### 2.2 Governance, observability, and structure in AI-enabled systems

AI-enabled systems increasingly require observability, control, and explicit runtime constraints [15, 16]. Work on software engineering for AI systems emphasizes inspectability and lifecycle-aware control, while structured-output studies show that schema-constrained generation can reduce ambiguity in downstream use [17]. These concerns motivate a stronger governance layer in long-horizon knowledge synthesis.

These concerns motivate a stronger governance layer in long-horizon knowledge synthesis. In the present architecture, governance is not a post-hoc quality filter but a runtime mechanism that merges heterogeneous evaluations, updates epistemic state, reallocates attention budget, and records evidence for later inspection [18].

## 2.3 Knowledge sharing, memory compression, and artifactized synthesis

Knowledge-oriented applications of LLMs increasingly rely on explicit external artifacts rather than transient chat turns . Long-context work further shows that compression and structured state can stabilize extended reasoning windows [18, 19]. The present architecture extends those ideas by treating raw code, formatted reports, section drafts, checkpoints, segment summaries, and per-round traces as structured explicit knowledge artifacts.

The present architecture extends those ideas by treating raw code, formatted reports, section drafts, checkpoints, segment summaries, and per-round traces as structured explicit knowledge artifacts. The objective is not merely to emit outputs, but to keep knowledge-state progression bounded, inspectable, and recoverable across long-horizon execution[20, 21].

*Table 1. Positioning of the proposed architecture relative to nearby lines of work.*

| Direction | Representative work | Primary focus | What remains underexplored |
|---|---|---|---|
| Multi-agent collaboration frameworks | CAMEL, MetaGPT, AutoGen, AgentVerse | Role specialization, collaboration patterns, framework generality | Explicit knowledge-state tracking, recoverable continuity, and evidence-bearing governance |
| Iterative reasoning and feedback | ReAct, Self-Refine | Stepwise reasoning and self-correction | Bounded knowledge-state progression and inspectable artifact chains |
| Software-engineering task benchmarks | SWE-bench, software-agent surveys | Task completion on engineering problems | Governance-intensive synthesis rather than schema-only output validity |
| Present study | Cognitive state-driven MAS architecture | Explicit state, consensus governance, recoverable continuity | System-level trade-off among governance benefit, continuity, and overhead |

## 3. Research objectives and questions

The evaluation is organized around knowledge-systems questions rather than benchmark-style claims of answer optimality or generic workflow throughput.

RQ1: Can the architecture maintain explicit epistemic progression and produce complete evidence-bearing knowledge artifacts across heterogeneous long-horizon synthesis tasks?

RQ2: Which mechanisms are most central to preserving knowledge-state continuity under interruption, coordination friction, and bounded context?

RQ3: What do artifact-integrity and protocol-level code-oriented signals reveal about the structural quality of generated explicit knowledge artifacts?

## 4. Proposed ADEMA runtime architecture

### 4.1 Architecture overview

The architecture starts from a task specification, role map, and budget profile. Specialized agents contribute in serial relay order, after which a dual-evaluator controller updates direction, milestones, summaries, and recovery-relevant state. The runtime therefore couples generation with explicit state update rather than treating multi-agent dialogue as unstructured accumulation. Figure 1 summarizes the ADEMA framework and shows how initialization, serial relay, persistence, synthesis, artifact assembly, and auditability are separated as software responsibilities.

Unlike free-form message accumulation, this formulation externalizes crucial latent variables: what is already known, what remains tentative, what was invalidated, which milestones were completed, and which artifacts are already stable enough to survive interruption or reviewer inspection. Taken together, these choices form a reusable architectural blueprint for governable long-running knowledge systems rather than a task-specific prompt script.

*Algorithm 1. Recovery-aware orchestration loop for explicit knowledge synthesis.*

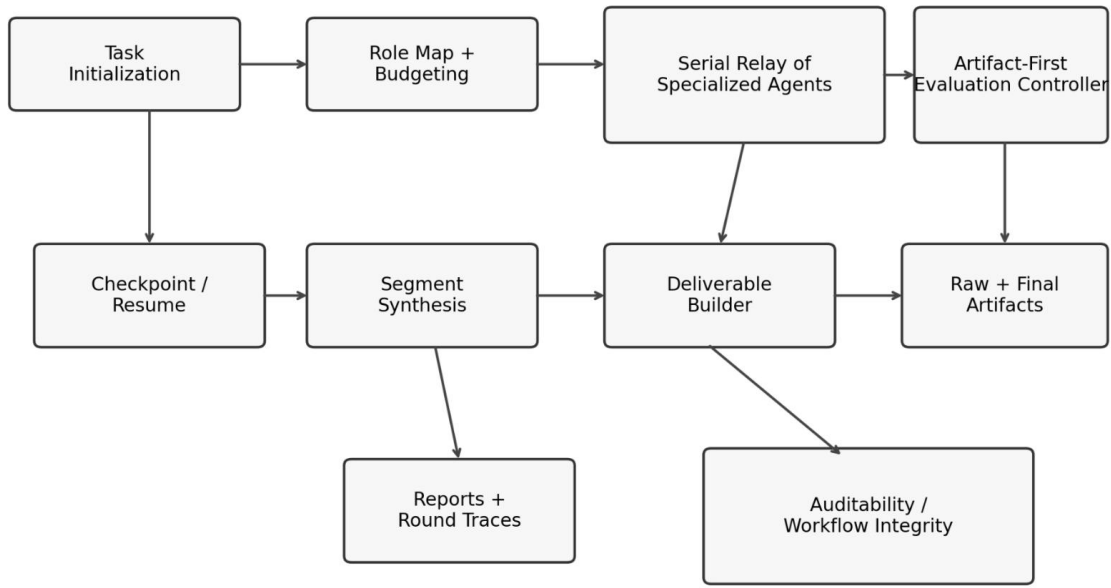

Figure 1. ADEMA framework with separated responsibilities for initialization, evaluation, recovery, compression, artifact assembly, and workflow integrity.

```
Input: task specification q, role map R, budget B, prior state S (optional)
1:  initialize run context, milestones, and deliverable builder
2:  if checkpoint exists and recovery is enabled then reload S
3:  while termination criteria are not met do
4:      dispatch the next specialized agent under the current route and budget
5:      collect candidate contribution and append round trace
6:      evaluate progress with the artifact-first controller
7:      update milestones, summaries, and next direction
8:      if interruption or crash occurs then persist checkpoint and exit safely
9:      if context budget becomes tight then perform segment synthesis
10:     update raw/final deliverable state and audit logs
11: end while
12: finalize report, raw artifacts, and formatted deliverables
```

## 4.2 Explicit epistemic state management

The source code maintains a hypothesis table with discrete states such as proposed, validating, proven, and disproven. Together with milestone progress and per-round summaries, this table acts as an explicit epistemic bookkeeping layer over the evolving synthesis process.

This mechanism is central to the KBS positioning because it converts otherwise implicit reasoning into inspectable symbolic state. The architecture therefore supports epistemic bookkeeping rather than free-form accumulation, enabling bounded knowledge-state progression, traceable state transitions, and drift control under long-horizon execution.

## 4.3 Heterogeneous dual-agent consensus governance

Round-level control signals are produced by a primary evaluator and a secondary evaluator. Their score vectors are merged through a weighted combination, after which agreement-sensitive control decisions are applied to milestone updates, route correction, and output validation.

Because the evaluators rely on heterogeneous backend families, the mechanism is better understood as cross-model consensus governance than as repeated self-judgment by a single model family. Its role is to strengthen knowledge-state update discipline, damp single-model bias, and provide supporting governance for long-horizon synthesis rather than act as a universal completion prerequisite.

## 4.4 Dynamic reputation and adaptive resource allocation

Each scientist agent maintains a reputation profile based on innovation events and total evaluated participation. The runtime converts this profile into a bounded token quota so that higher-value contributors gain more speaking budget while persistently weak contributors can be down-weighted or reallocated.

This mechanism serves two knowledge-systems functions. First, it allocates scarce context and attention budget adaptively rather than uniformly. Second, it provides a self-healing path when an agent enters a sustained low-value regime, thereby reducing the chance that cognitive drift is amplified by repeated weak contributions.

## 4.5 Long-term memory condensation and stateful persistence

The implementation periodically synthesizes earlier rounds into segment summaries once the history threshold is reached. This memory condensation reduces prompt growth while preserving the highest-value continuity cues. In parallel, checkpoint persistence serializes a recovery-relevant approximation of the current epistemic state so that interrupted runs can re-enter without rebuilding the entire chain from scratch.

These two mechanisms address different but related sources of instability. Segment synthesis mitigates attention dilution under long context. Checkpoint persistence mitigates catastrophic state loss under interruption. Within this pair, persistence is the stronger recoverability mechanism, whereas memory condensation primarily serves continuity preservation under bounded context.

### 4.6 Knowledge artifact assembly

The final stage is task-specific but conceptually unified. Code tasks produce raw code, formatted code, and companion documentation. Literature tasks produce raw section-level text and formatted review documents. Structured exploratory tasks produce explicit mechanism-oriented reports. In KBS terms, this layer functions as knowledge-artifact assembly: it materializes evidence-bearing outputs that can be audited, handed off, or resumed after interruption[22].

Table 2 summarizes the main architectural mechanisms, and Table 3 operationalizes the explicit knowledge-state and evidence-bearing capabilities used throughout the review package.

*Table 2. Main mechanisms and their roles in long-horizon knowledge synthesis.*

| Mechanism | Description | Knowledge-systems role |
|---|---|---|
| External configuration and snapshots | YAML-backed parameters plus per-run snapshots | Reproducibility and tunable state discipline |
| Dual logging | Console and file logging during runtime | Inspectability and post-hoc auditability |
| Checkpoint/resume | Persistent state saving and reload paths | Recoverable continuity under interruption |
| Dual evaluation | Primary and secondary evaluators with merged control outputs | Consensus governance and route correction |
| Segment synthesis | Periodic compression of prior progress | Bounded context and continuity preservation |
| Deliverable builder | Task-specific raw/final artifact assembly | Evidence-bearing artifact assembly |
| Report generation | Scenario report plus round-level traces | Reviewable knowledge-state evidence chain |

*Table 3. Explicit knowledge-state and evidence-bearing capabilities used in the review package.*

| Capability | Operational meaning | Evidence source |
|---|---|---|
| checkpoint persistence | Persists controller state and task-directory snapshots for continued execution after interruption. | checkpoint files and reload path |
| round traces | Records per-round agent contributions and evaluator states for post-hoc review. | round_*.json artifacts |
| dual logging | Writes both console and file logs during runtime. | run log plus report references |
| segment synthesis | Compresses earlier progress while preserving continuity cues for long runs. | segment synthesis outputs |
| raw/final deliverables | Emits raw artifacts and formatted handoff-ready outputs. | raw_* and final_* artifacts |
| watchdog-assisted continuity | Supports monitored continuation in interruption-sensitive | resume showcase evidence chain |

| Capability | Operational meaning | Evidence source |
|---|---|---|
| | execution. | |
| evaluator-mediated control | Uses merged evaluator feedback to update direction and milestones. | controller updates and round summaries |
| evidence bundling | Packages reports, logs, traces, checkpoints, and deliverables as one reviewable chain. | artifact inventory and report bundle |

Table 3 operationalizes the architecture's explicit capabilities rather than merely naming them. Table 3b then links implementation mechanisms to their principal evidence blocks, and Table 3c retains only the capability subset most essential to the main-text KBS argument. This is important for the KBS framing because the manuscript's strongest claim is not that ADEMA is a faster runtime, but that hidden long-horizon state is made observable through explicit, evidence-bearing capabilities. Checkpoint persistence is the clearest completion-critical mechanism under interruption; dual evaluation, segment synthesis, and replacement are supporting control mechanisms whose primary role is trajectory shaping rather than universal binary necessity.

To make the code-level contribution auditable, the manuscript now makes the implementation layer explicit. Table 3b reorganizes the reviewed source code into eight runtime mechanisms and maps each mechanism to the main evidence block that most directly exercises it. This table shows that the contribution is not a loose collection of prompts, but a stateful software runtime whose control semantics are implemented, inspectable, and empirically grounded.

### 4.7 Implemented runtime mechanisms and empirical mapping

Table 3b. Implemented ADEMA runtime mechanisms and their main evidence blocks.

Table 3c. Simplified capability checklist retained in the main text.

| Capability | Operational meaning | Evidence source |
|---|---|---|
| checkpoint persistence | Persists controller state and task-directory snapshots for continued execution after interruption. | checkpoint files and reload path |
| round traces | Records per-round agent contributions and evaluator states for post-hoc review. | round_*.json artifacts |
| dual logging | Writes both console and file logs during runtime. | run log plus report references |
| segment synthesis | Compresses earlier progress while preserving continuity cues for long runs. | segment synthesis outputs |
| raw/final deliverables | Emits raw artifacts and formatted handoff-ready outputs. | raw_* and final_* artifacts |
| watchdog-assisted continuity | Supports monitored continuation in interruption-sensitive execution. | resume showcase evidence chain |
| evaluator-mediated control | Uses merged evaluator feedback to update direction and milestones. | controller updates and round summaries |
| evidence bundling | Packages reports, logs, traces, checkpoints, and deliverables as one reviewable chain. | artifact inventory and report bundle |

The full 12-item checklist, including detection rules and trace-to-artifact mapping, is provided in the supplementary material. The condensed main-text version is retained here because it operationalizes the architecture rather than merely narrating it.

| Runtime mechanism | Code-level realization | Primary empirical evidence |
|---|---|---|
| Explicit epistemic state | Hypothesis table, milestone bookkeeping, round summaries, and per-round state updates | Showcase evidence; Tables 3 and 4 |
| Dual evaluation | Primary and secondary judges with weighted score merging and strict validity fusion | B-code ranking block; mechanism matrix; dynamic-governance trace |
| Adaptive task-mode switching | Mismatch detection, reconfiguration of score dimensions, milestones, and agent roles | Runtime architecture section; trace-level behavior in long-horizon cases |
| Dynamic replacement | Low-efficiency detection, double-confirmed replacement, and replacement-event logging | Targeted micro-ablation of dynamic governance |
| Checkpoint-resumable persistence | Task-directory checkpoints, reload path, and interruption-sensitive restart logic | Interruption-sensitive rerun; repaired Experiment A block |
| Segment-level synthesis | Periodic segment synthesis, compressed memory, and segment artifacts | Mechanism sensitivity and long-horizon compression evidence |
| Artifact-first deliverable builder | Task-specific code, literature, and structured-output builders | Structural integrity proxies; FACP; showcase artifacts |
| Final validation and fallback | Syntax/compilation checks, output-valid flags, and fallback to safer raw artifacts | Final deliverable validity audit; quality-proxy evidence |

## 5. Evaluation methodology

### 5.1 Evidence packages and tasks

The empirical basis of this submission is organized as an integrated experimental study centered on knowledge-state questions. The first package is a four-scenario showcase covering code synthesis, literature synthesis, interruption-sensitive resume, and a high-constraint structured reasoning scenario. The second package is a fixed 60-run matrix that combines Full MAS, a single-model baseline, and targeted mechanism ablations. The full experimental matrix specification is provided in the supplementary material, while the main text retains only the condensed comparisons needed for headline interpretation [23, 24].

A smaller seven-case black-box campaign and a unified benchmark package are retained as supplementary robustness layers rather than alternative headline result sources. Their role is to broaden the evidence base around objective artifact scores, representative edge cases, and row-level execution heterogeneity while keeping the main manuscript centered on the four-scenario showcase and the fixed 60-run matrix.

### 5.2 Measures and interpretation

The evaluation uses a main-and-supporting indicator structure. Artifact-valid completion is treated as the primary system-level outcome because the architecture is designed to preserve explicit state, evidence-bearing artifacts, and resumable continuity rather than only process exit status. Runtime, token cost, success rate, and interruption-sensitive continuity are reported as system-level operational properties. Evidence richness, artifact inventory, and final artifact-chain completeness act as supporting architecture-level signals.

To complement binary completion, the study reports lightweight structural and artifact-oriented indicators, including reference-bearing outputs, explicit mechanism-bearing structured reports, and recoverable raw/final artifact chains. These indicators are interpreted as supporting evidence for inspectable knowledge progression rather than as gold-standard semantic truth measures.

## 5.3 Evaluation-layer organization

The evaluation is organized into two layers. The first is an archived structural audit layer covering innovation, logic, focus, and completeness. This layer underlies the ranking evidence used throughout the current manuscript. The second is a complementary semantic layer covering factual consistency, instruction adherence, and semantic coherence.

In the present submission, the semantic layer is retained as an explicit methodological extension rather than as a new headline score block. Full numeric population of these additional dimensions would require a fresh rerun under the same artifact set. This separation keeps the current evidence package honest while still clarifying the content-level quality dimensions that matter for knowledge-oriented artifacts.

## 6. Results

Two primary empirical signals organize the result section. Primary empirical signal 1 is that checkpoint persistence is the only mechanism that creates a visible completion boundary under interruption. Primary empirical signal 2 is that the code-oriented benchmark is the clearest quality-sensitive mechanism block. All remaining evidence layers are interpreted as supporting signals for governance discipline, artifact-chain completeness, and cost-quality behavior rather than as competing headline claims.

### 6.1 Showcase evidence of explicit knowledge-state continuity

The showcase package supports three direct claims. First, the implementation exposes the planned mechanisms as an integrated architecture rather than as isolated features. Second, all four heterogeneous scenarios completed successfully. Third, each scenario produced a full evidence chain containing reports, traces, checkpoints, segment summaries, and raw/final artifacts. These results support the claim that explicit state and evidence-bearing outputs can be maintained across heterogeneous synthesis tasks.

The interruption-sensitive resume showcase is the single most informative case because it required two attempts and one watchdog-assisted continuation yet still returned a complete artifact chain. In knowledge-systems terms, the case demonstrates stateful epistemic persistence rather than merely process survival.

*Table 4. Four-scenario showcase results and evidence-chain patterns.*

| Scenario | Success | Attempts | Resumed | Watchdog | Overall | Key artifact pattern |
|---|---|---|---|---|---|---|
| Code Showcase | Yes | 1 | No | 0 | 9.40 | Report + raw/final code + segment + checkpoint + round traces |
| Literature Showcase | Yes | 1 | No | 0 | 9.40 | Report + raw/final literature + segment + checkpoint + round traces |
| Resume Showcase | Yes | 2 | Yes | 1 | 9.76 | Recovered run with full evidence chain preserved |
| LATAM Compliance-Control Paradox | Yes | 1 | No | 0 | 9.25 | Structured report with full process evidence |

### 6.2 Mechanism sensitivity in the 60-run matrix

The repeated matrix produced 60 runs, of which 59 were artifact-valid under the finalized artifact-first controller. Full MAS, the single-model baseline, the dual-evaluation ablation, and the segment-synthesis ablation each completed all 12 scenario-seed combinations successfully. The only invalid run occurred when checkpoint/resume was removed, and it occurred in the interruption-sensitive resume condition. This remains the clearest mechanism-level result in the study.

Table 5 shows that Full MAS is substantially slower than the single-model baseline. In knowledge-state terms, that difference should be read as governance overhead: the additional runtime reflects the cost of explicit bookkeeping, heterogeneous evaluation, persistence, and artifact assembly. Figure 2 visualizes artifact-valid success rate by configuration, and Figure 3 contrasts mean runtime.

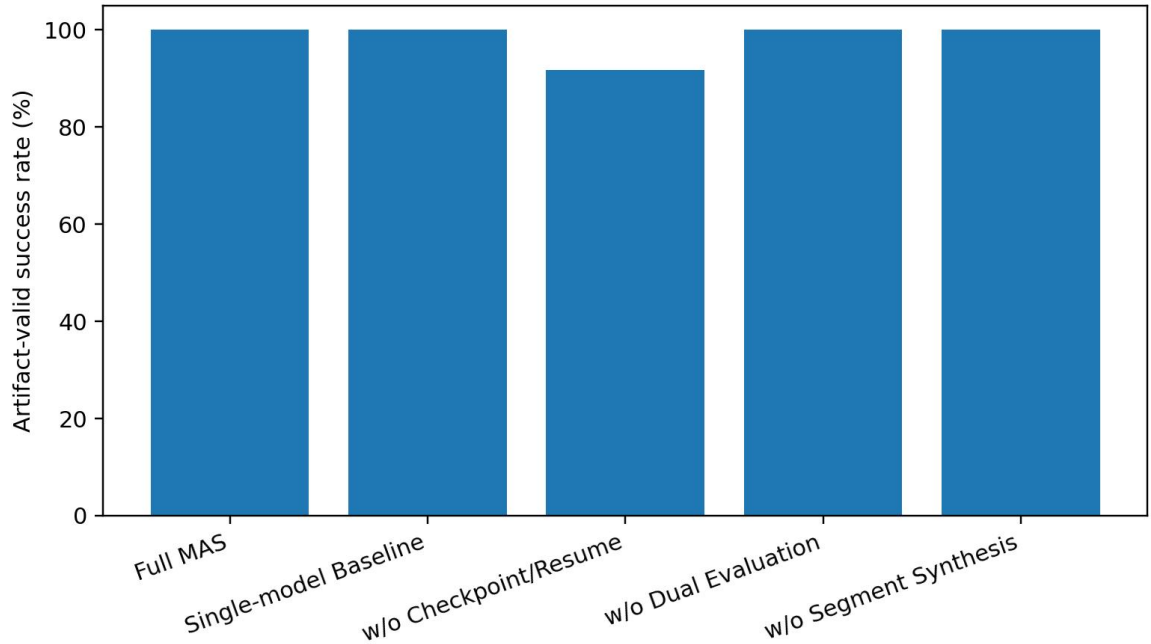

Figure 2. Artifact-valid success rate by configuration in the fixed 60-run matrix.

*Table 5. Configuration-level summary from the fixed matrix.*

| Configuration | Success rate | Mean runtime | Median runtime | Reports/run | Checkpoints/run | Raw/run | Final/run |
| --- | --- | --- | --- | --- | --- | --- | --- |
| Full MAS | 100.0% | 916.0s | 773.8s | 1.0 | 1.0 | 1.5 | 1.0 |
| Single-model Baseline | 100.0% | 292.0s | 296.0s | 1.0 | 1.0 | 1.5 | 1.0 |
| w/o Checkpoint/Resume | 91.7% | 877.6s | 819.8s | 1.0 | 0.0 | 1.3 | 0.9 |
| w/o Dual Evaluation | 100.0% | 1118.5s | 1056.7s | 1.0 | 1.0 | 1.5 | 1.0 |
| w/o Segment Synthesis | 100.0% | 793.5s | 695.5s | 1.0 | 1.0 | 1.5 | 1.0 |

## 6.3 Heterogeneous consensus, memory condensation, and adaptive control

Removing dual evaluation did not reduce completion success in the reported matrix. The KBS interpretation is therefore specific: heterogeneous consensus should be understood primarily as a governance-intensive mechanism. At the present workload scale, its effect is layered rather than binary. It shapes correction, confidence, and control discipline even when completion remains possible without it.

Removing segment synthesis likewise left completion success unchanged in the 60-run matrix. This indicates that memory condensation is best interpreted here as a continuity-preserving and compression-oriented mechanism. Its main contribution is to stabilize long-history progression and reduce context growth, and its effect at the tested workload scale is layered rather than binary.

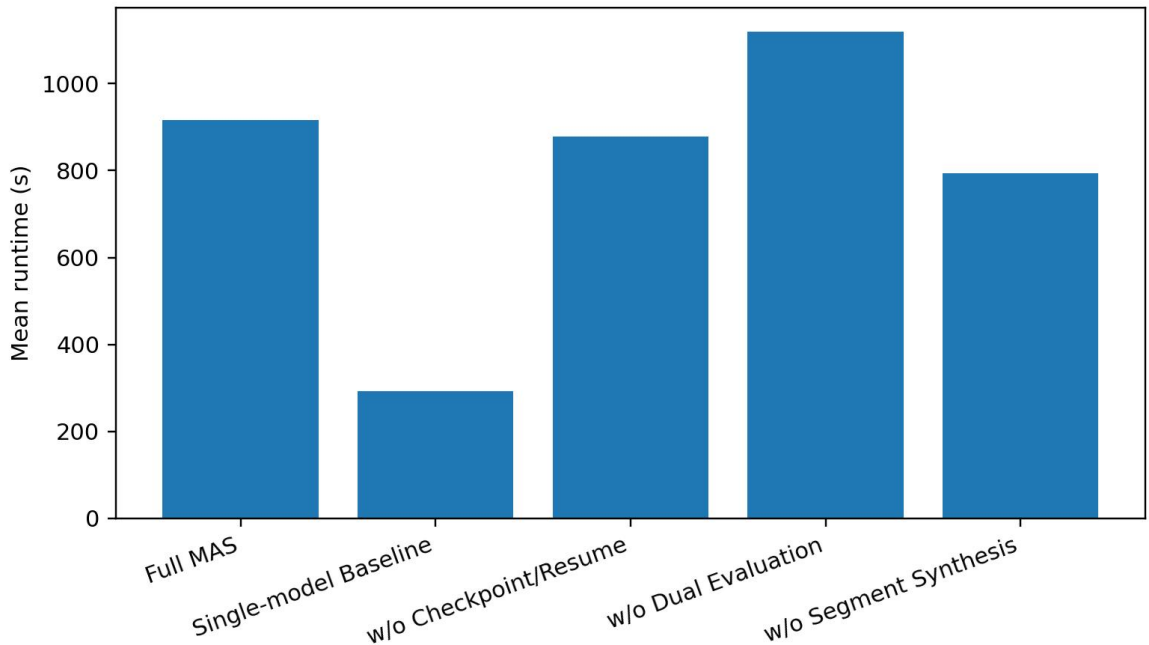

Figure 3. Mean runtime by configuration in the fixed matrix.

Table 6. Scenario-specific result of removing checkpoint/resume.

| Scenario | Successes | Runs | Success rate | Interpretation |
| --- | --- | --- | --- | --- |
| code | 3 | 3 | 100.0% | No immediate interruption sensitivity observed |
| literature | 3 | 3 | 100.0% | Sectional accumulation still completed |
| resume | 2 | 3 | 66.7% | Only scenario showing a clear penalty from removing persistence |
| latam | 3 | 3 | 100.0% | Exploratory structured reporting still completed |

## 6.4 Structural integrity and artifact-chain evidence

The added artifact-oriented evidence indicates that successful runs were not merely process-complete. Across the finalized evidence package, the most defensible structural signal is not a weak low-level proxy but the presence of reviewable raw/final artifact chains, interruption-recovery traces, and task-complete evidence bundles. The interruption-sensitive rerun, in particular, shows that ADEMA can restore a usable evidence chain after deliberate disruption, including reports, checkpoints, round traces, and raw/final artifacts.

Table 7. Structural artifact-chain evidence and interruption-sensitive continuity results.

| Added evidence | Scope | Key result | Interpretation |
| --- | --- | --- | --- |
| Artifact-chain evidence update | Finalized evidence package | Reviewable raw/final artifact chains and evidence-bearing outputs were preserved across the main successful runs. | Supports structural inspectability without relying on a weak syntax-proxy headline. |

| Added evidence | Scope | Key result | Interpretation |
|---|---|---|---|
| Interruption-sensitive rerun | Current system, code workflow | First execution interrupted; resumed execution restored reports, checkpoints, round traces, and raw/final artifacts. | Supports recoverable continuity and evidence-chain restoration after interruption. |

### 6.5 Supplementary black-box robustness signal

The supplementary black-box campaign is retained only as a robustness layer rather than as a headline validity block. Its main value is qualitative: the cases show that ADEMA can sustain resumed execution, automatic mode adaptation, and long-horizon compression while still producing reviewable artifacts under heterogeneous conditions. In the revised KBS framing, these cases broaden the evidence base but do not redefine the paper's strongest claims.

Figure 4. Objective artifact score in the seven-case supplementary black-box campaign.

### 6.6 Targeted micro-ablation of dynamic governance

A targeted four-run micro-ablation isolated replacement-based dynamic governance under a low-cost exploration task. Group A disabled replacement, whereas Group B enabled threshold-based replacement. All four runs formed valid outputs, so this small experiment should not be read as a completion-boundary test. Its value lies in trajectory shaping and cost-quality interpretation.

Table 8. Targeted micro-ablation summary for dynamic governance.

Table 8 shows a small but directionally favorable governance-efficiency signal: the dynamic-governance condition achieved a slightly higher final EMA while using fewer tokens and lower mean cost. Only one governance-enabled run actually triggered three replacements, but each trigger was followed by EMA recovery. The present evidence therefore supports trajectory shaping, control discipline, and cost-quality interpretation for dynamic governance, not a universal necessity claim.

| Group | Runs | Mean final EMA | Mean delivery | Mean tokens | Mean cost (USD) |
|---|---|---|---|---|---|
| A: replacement disabled | 2 | 8.36 | 94.0% | 65,741 | $0.110631 |
| B: dynamic governance enabled | 2 | 8.40 | 93.5% | 61,320 | $0.093249 |

### 6.7 Final artifact completeness proxy

To complement the micro-ablation, the same four runs were re-examined using a Final Artifact Completeness Proxy (FACP). The proxy scores four structural dimensions of the delivered artifact chain: final-file presence, raw/final pair presence, required-section coverage, and trace integrity. This layer asks not whether a run simply finished, but whether it produced a complete evidence-bearing artifact chain.

Table 9. Final Artifact Completeness Proxy (FACP) by condition.

| Group | Runs | Final files | Raw/final pairs | Required-section coverage | Mean FACP |
|---|---|---|---|---|---|
| A: replacement disabled | 2 | 2/2 | 2/2 | 14/14 | 1.00 |
| B: dynamic governance enabled | 2 | 2/2 | 2/2 | 14/14 | 1.00 |

Table 9 indicates that both conditions achieved full artifact-chain completeness under this proxy. In other words, the targeted micro-ablation differentiates runtime-cost and trajectory-correction behavior more clearly than final deliverable formation. This is precisely why the manuscript treats artifact-chain completeness as a supporting structural layer rather than as the single headline result.

The implication is still important: ADEMA is not designed as a message-only workflow. It is designed to carry explicit evidence-bearing artifacts across rounds, interruptions, and formatting stages. Full FACP in both conditions therefore strengthens the claim that explicit artifact progression is architecturally stable even when control mechanisms are varied.

### 6.8 Forced interruption and recovery evidence

A direct forced-interruption test further supports the persistence claim. Under the normal baseline, the interruption-sensitive workflow reached a final EMA of 8.64 with 89.6% delivery completion. Under an interrupted-and-resumed condition, the run still recovered to a final EMA of 8.45 with 89.0% delivery completion while consuming markedly fewer total tokens. This is not the strongest quality-sensitive result in the paper, but it is a clean recoverability demonstration and is consistent with the fixed-matrix finding that persistence is the only mechanism creating a visible completion boundary under interruption.

Table 10. Forced interruption and recovery test results.

The value of this forced-recovery evidence is interpretive rather than competitive: it shows that recoverable continuity can preserve task-level progress and a usable evidence chain even when execution is deliberately broken and resumed.

### 6.9 Ranked protocol-level and horizontal baseline comparison

Code-oriented tasks are treated here as a strict artifact-validity stress case rather than as the sole application identity of the architecture. Compared with softer text-only outputs, code artifacts expose state drift, coordination failure, and deliverable invalidity earlier because structural defects become more immediately observable. For that reason, the protocol-level B-code block is used as the clearest quality-sensitive mechanism test in the present evidence package.

| Rank | Evidence block | Representative result | Reference comparison | Interpretation | Executable files/run | Documentation/run | Resume/state retention |
|---|---|---|---|---|---|---|---|
| 1 | Experiment B-code | FullMethod scored 8.00 and 7.93 | Tied Gemini iter10 at the top and remained above author baseline (7.10) | Strongest quality-sensitive mechanism block for code-centric innovation. | 1.0 | 1.0 | Recoverable by design |
| 2 | Experiment A (interruption) | Checkpoint-on runs scored 7.08–7.53 | Above author baseline (6.50–6.90) and close to | Recoverability preserved output quality under | 1.0 | 1.0 | Resumed chain preserved |
| | | | Gemini iter5 (7.00) | interruption-sensitive execution. | | | |
| 3 | Experiment B-analysis | FullMethod reached 7.60 and 7.30 | Above author baseline but below Gemini iter10 (8.30) | Supporting rather than headline evidence because separations are smoother. | 0.0 | 0.0 | Not applicable |
| 4 | AutoGen heterogeneous baseline | Normal: 130.73 s, 4.3 interactions/run; crash: 0/3 resumed | Much faster, but with 0 checkpoints, 0 executable files, and 0 documentation | Paradigm-difference reference showing the contrast between stateless conversation and stateful orchestration, not a fair engineering race. | 0.0 | 0.0 | 0/3 resumed; 0% state retention |

Table 11. Condensed protocol-level evidence with the AutoGen horizontal baseline included as a paradigm-difference reference.

The AutoGen comparison is not intended as a framework-fair engineering race. Its purpose is to contrast a stateless conversational paradigm with a stateful orchestration paradigm under long-horizon, artifact-bearing tasks. In that sense, AutoGen serves as a paradigm-difference reference: it is faster, but it does not preserve checkpoints, executable artifact chains, companion documentation, or resumable state under crash stress.

| Evidence block | Representative result | Reference comparison | Interpretation | Role in main narrative |
|---|---|---|---|---|
| Experiment A (interruption) | Checkpoint-on runs scored 7.08-7.53 | Above author baseline (6.50-6.90) and close to Gemini iter5 (7.00) | Recoverability preserved output quality under interruption-sensitive execution. | Supporting recoverability evidence |
| Experiment B-code | FullMethod scored 8.00 and 7.93 | Tied Gemini iter10 at the top and remained above author baseline (7.10) | Strongest quality-sensitive mechanism block for code-centric innovation. | Primary headline evidence |
| B-code ablations | NoSegmentSync and NoEvaluator-NoSegmentSync dropped to 7.63/7.40/7.05/6.92 | Below FullMethod and below or near top references | Evaluator-mediated control and segment synchronization both contribute positively. | Mechanism-sensitivity evidence |
| Experiment B-analysis | FullMethod reached 7.60 and 7.30 | Above author baseline but below Gemini iter10 (8.30) | Supporting rather than headline evidence because separations are | Supporting evidence only |

| | | | smoother. | |
|---|---|---|---|---|
| | | | | |

## 7. Discussion

The combined evidence supports a knowledge-state orchestration interpretation rather than a generic multi-agent superiority claim. ADEMA's strongest contribution is that it makes long-horizon knowledge synthesis governable: hypotheses become explicit state, milestones become managed progress signals, artifacts become evidence-bearing runtime objects, and interruption becomes recoverable continuity rather than silent state loss.

One implication concerns mechanism hierarchy. The present evidence supports necessity most strongly for checkpoint persistence. By contrast, dual evaluation, segment synthesis, and dynamic governance are better interpreted here as supporting control mechanisms: they shape trajectory discipline, cost-quality behavior, and artifact progression, but the current package does not justify universal necessity claims for them at the tested workload sizes[25].

A further implication concerns governance overhead. The fixed matrix shows that governed synthesis is not free: richer bookkeeping, heterogeneous consensus, persistence, and artifact assembly all impose runtime cost relative to a simpler baseline. Yet this extra cost is not noise around the experiment; it is part of the result. The architecture trades raw speed for controlled progression, evidence richness, and recoverable continuity. The data therefore support a cost-of-governed-synthesis reading rather than a deficit-only interpretation of slower runtime. The AutoGen baseline should therefore be read as paradigm-difference evidence—stateless conversational coordination versus stateful orchestration under long-horizon tasks—rather than as a claim that the two implementations constitute a like-for-like engineering contest.

A third implication is architectural. The five interacting mechanisms—explicit state management, heterogeneous consensus governance, dynamic reputation allocation, stateful persistence with memory condensation, and knowledge-artifact assembly—form a reusable reference architecture for long-horizon knowledge synthesis. The present study does not claim universality, but it does provide a concrete blueprint for systems that must keep knowledge progression inspectable, restartable, and reviewable under real execution constraints.

A targeted micro-ablation further sharpens this interpretation. Under a small four-run controlled comparison, dynamic governance did not create a new completion boundary, because both conditions still formed valid deliverables. What changed instead was the efficiency–trajectory combination: the governed condition showed slightly better final EMA with lower mean token cost, and one governed run exhibited replacement-triggered EMA recovery. Read together with the fixed matrix, this supports a graceful-degradation view in which adaptive governance matters most as runtime correction and cost-aware control rather than as a universal on/off determinant of completion.

## 8. Threats to validity

First, the evaluation emphasizes explicit knowledge-state progression, recoverable continuity, and artifact integrity rather than external gold-standard semantic correctness tasks. This improves alignment with the architecture under study, but limits direct comparability with benchmark-style evaluations[26].

Second, the quality proxies are intentionally lightweight. Reference presence, mechanism presence, and artifact-chain completeness are useful structural indicators, yet they do not by themselves establish factual correctness, domain optimality, or semantic sufficiency. The manuscript should therefore be read as a knowledge-state architecture study with bounded outcome-level quality evidence.

Third, the repeated matrix probes only a limited set of ablations. It supports a strong interpretation for checkpoint persistence, but more layered interpretation for dual evaluation, segment synthesis, and dynamic governance. The current package supports trajectory shaping, control discipline, and cost-quality interpretation for these mechanisms, not universal necessity claims.

Fourth, the implementation under study is one concrete architecture rather than a broad family of systems. The conclusions should therefore be interpreted as mechanism-local evidence and reference-architecture insight rather than as a universal law for all multi-agent systems. The AutoGen comparison is likewise interpretive rather than tournament-style: it is included to show how a stateless conversational paradigm differs from a stateful orchestration paradigm under the same long-horizon task family, not to claim a perfectly matched framework-fair engineering race.

## 9. Conclusion

This paper presents ADEMA as a knowledge-state orchestration architecture for governable long-horizon knowledge synthesis. The central result is not that multi-agent orchestration is universally faster or semantically superior. It is that explicit epistemic progression, evidence-bearing artifact progression, and recoverable continuity can be organized as concrete architectural commitments rather than left implicit in free-form interaction.

The strongest mechanism-level finding concerns persistence. Removing checkpoint/resume produced the only invalid run in the fixed matrix and did so exactly where architectural intuition predicts: the interruption-sensitive resume condition. The clearest quality-sensitive mechanism block, however, appears in the code-oriented benchmark, where FullMethod most clearly separates itself from ablated variants and the author baseline.

For Knowledge-Based Systems, the broader takeaway is that long-horizon LLM collaboration should be designed around explicit knowledge state, selective governance, consensus-aware control, and artifactized progression rather than around unconstrained conversation alone. Immediate next steps include stronger human judgment of artifact usefulness, harsher interruption regimes, and matched framework comparisons under the same evidence-bearing output contract.

### Appendix A. Mechanism-role interpretation

For review convenience, the compact mechanism-role interpretation used during manuscript refinement is retained here as an appendix-level summary. The fixed matrix and reviewer-facing evidence chain support the interpretation reported in Table A1.

*Table A1. Compact interpretation of mechanism roles under the finalized evidence package.*

| Mechanism | Evidence from exp34 | Interpretation |
|---|---|---|
| Checkpoint/Resume | Only removal condition causing the sole invalid run (resume scenario: 2/3 successes without persistence). | Clearest completion-critical mechanism under interruption. |
| Dual Evaluation | 12/12 successful runs even when the secondary | Governance-intensive mechanism |

| Mechanism | Evidence from exp34 | Interpretation |
|---|---|---|
| | evaluator is removed. | with layered rather than binary effects at current workload scale. |
| Segment Synthesis | 12/12 successful runs even when intermediate synthesis is removed. | Continuity-preserving and compression-oriented mechanism; layered effect under current workload scale. |
| Multi-agent orchestration | Full MAS remains substantially slower than the single-model baseline despite equivalent completion success. | Reflects epistemic governance cost paid for inspectability, recoverability, and evidence richness. |

## Appendix B. Reviewer-facing execution checklist

To support editor and reviewer inspection, Table B1 condenses the evidence-bearing execution checklist used to verify that ADEMA is being assessed as a knowledge-state orchestration architecture rather than as a free-form conversation script. The full expanded checklist and row-level trace mapping are provided in the supplementary material.

Table B1. Execution checklist used for reviewer-facing package verification.

| Checkpoint item | Operational verification question | Main evidence location |
|---|---|---|
| Explicit epistemic state | Are hypotheses, milestones, and summaries externalized as inspectable runtime state? | Section 4.2; Tables 3 and 3c |
| Evaluator-mediated governance | Is direction correction governed by a heterogeneous controller rather than free-form chat continuation? | Section 4.3; Tables 2 and 3b |
| Adaptive role governance | Is role-level contribution discipline visible through dynamic reputation or replacement logic? | Section 4.4; Table 8 |
| Checkpoint persistence | Can interrupted execution resume from persisted state rather than restart from scratch? | Sections 4.5, 6.2, 6.8; Table 10 |
| Memory condensation | Is long-horizon continuity preserved through segment-level synthesis? | Sections 4.5 and 6.3 |
| Artifact-first assembly | Does the system emit raw/final paired artifacts rather than text-only answers? | Section 4.6; Table 9 |
| Evidence bundling | Are reports, traces, checkpoints, and deliverables preserved as one reviewable chain? | Tables 3c, 7, 9; Supplementary package |
| Horizontal baseline contract | Does the lightweight baseline satisfy the same artifact and recovery contract? | Section 6.9; Table 11a |

## Appendix C. Submission package map

Table C1 maps the submission package to the reviewer-facing claims in the main manuscript and explains the purpose and usage of the newly added materials. The framework figure clarifies responsibility separation in the architecture section; the expanded supplementary scoring tables are for interpretive support rather than primary ranking; the package map and execution checklist are practical navigation aids; and the confidential reviewer guide remains the route for source-level inspection of the complete runnable implementation.

Table C1. Submission-package file map and review purpose.

The package map is included to reduce ambiguity during peer review. It clarifies how each added material should be used: the main manuscript carries the headline claims, the supplementary PDF carries expanded non-code evidence, the ZIP carries machine-readable artifacts and source documents, and the confidential reviewer guide explains when source-level inspection becomes necessary.

| File | Role in package | Primary review purpose |
| --- | --- | --- |
| 01_Manuscript_KBS_Main.docx | Main article | Primary argument, methods, and results |
| 02_Cover_Letter_KBS.docx | Editorial letter | Submission positioning and scope statement |
| 03_Highlights_KBS.txt | Metadata support | Short contribution bullets for submission system |
| 04_Data_Availability_Statement.txt | Transparency statement | Public/private access boundary |
| 05_Declaration_of_Generative_AI_Use.txt | Policy statement | Required declaration for AI-assisted writing workflow |
| 06_Supplementary_Material.pdf | Non-code supplement | Extended tables, methods notes, and reviewer-facing summaries |
| 07_Supplementary_Code_and_Artifacts.zip | Artifact archive | Code, traces, figures, tables, and representative source materials |
| 08_Confidential_Reviewer_Guide.docx | Confidential guide | Path for editor/reviewer inspection of the runnable implementation |

## Acknowledgements

Not applicable.

## CRediT authorship contribution statement

Zhou Hanlin: Conceptualization, methodology, software, validation, investigation, formal analysis, writing - original draft, writing - review & editing.

Chan Huah Yong: Supervision, conceptualization, methodology, writing - review & editing.

## Declaration of competing interest

The authors declare that they have no known competing financial interests or personal relationships that could have appeared to influence the work reported in this paper.

## Data availability statement

Public supplementary materials accompanying this submission are reorganized into two reviewer-friendly files: a single Supplementary Material document and a single Source Code and Artifacts archive. Together they provide pseudocode, a redacted configuration template, reviewer-oriented evidence summaries, row-level tables, representative traces, figures, and English-language source-code and artifact samples sufficient for technical inspection. The complete runnable implementation remains available to the editor and reviewers for confidential peer-review purposes.